\documentclass[11pt]{article}

\usepackage[final]{acl}

\usepackage{times}
\usepackage{latexsym}

\usepackage[T1]{fontenc}

\usepackage[utf8]{inputenc}

\usepackage{microtype}

\usepackage{inconsolata}

\usepackage{graphicx}

\usepackage{amsmath}
\usepackage{amssymb}
\usepackage{mathtools}
\usepackage{amsthm}
\usepackage{booktabs}
\usepackage{colortbl}

\usepackage{amsmath,amsfonts,bm}

\def\eqref#1{equation~\ref{#1}}

\def\1{\bm{1}}

\DeclareMathAlphabet{\mathsfit}{\encodingdefault}{\sfdefault}{m}{sl}
\SetMathAlphabet{\mathsfit}{bold}{\encodingdefault}{\sfdefault}{bx}{n}

\definecolor{amethyst}{rgb}{0.6, 0.4, 0.8}
\definecolor{bestspeed}{rgb}{1, 0.88, 0.88}

\usepackage{algorithm}
\usepackage{algpseudocode} 
\usepackage[most]{tcolorbox}

\usepackage{amsmath} 
\usepackage{amsfonts}
\usepackage{amssymb} 

\usepackage{graphicx} 
\usepackage{subcaption}
\usepackage{pdfpages}
\usepackage{adjustbox} 
\usepackage{floatflt} 
\usepackage{wrapfig}
\usepackage{array}
\usepackage{tabularray}

\usepackage{multirow}
\usepackage{multicol}
\usepackage{booktabs}
\usepackage{placeins}
\usepackage{xcolor}

\usepackage{lipsum} 
\usepackage{tikz-cd}

\usepackage{comment}
\usepackage{blindtext}

\title{Strong Drafts Need Compact Memories: \\Long-Context Speculative Decoding with Compressed KV Cache}

\author{
  \textbf{Tong Yuan},
  \textbf{Chengxi Liao},
  \textbf{Zeyi Wen\thanks{Corresponding author.}} \\
  Data Science and Analytics Thrust, Information Hub, \\
  The Hong Kong University of Science and Technology (Guangzhou) \\
  \texttt{\{tyuan053, cliao118\}@connect.hkust-gz.edu.cn, wenzeyi@hkust-gz.edu.cn}
}

\begin{document}
\maketitle
\begin{abstract}
Long-context LLM applications such as document summarization and multi-turn agents require generation from prefixes spanning tens of thousands of tokens, making decoding latency a major bottleneck. Speculative decoding (SD) reduces latency without changing model outputs, but its speedup depends on both accepted draft tokens and draft-step latency: Lightweight drafts are fast but lack the capacity to capture long-range dependencies, whereas strong independent drafts recover acceptance but incur growing KV-access cost at long prefixes. We introduce memory-augmented drafting for long-context SD, equipping a strong independent draft with compressed draft-side KV memory: A lightweight adaptor constructs and incrementally updates this memory to retain distant information and exact recent context. The target verifier retains its full KV cache and applies the standard accept/reject rule, preserving SD's lossless guarantee. Experiments on Llama~3.1-8B and 70B targets at prefix lengths up to 32K show that our method reduces draft-side memory by over 70\%. It achieves speedups of up to 2.08$\times$ and 3.33$\times$, respectively, over autoregressive decoding.

\end{abstract}


\section{Introduction}

As LLM applications such as document summarization, online deep research, and multi-turn agents grow in scope and complexity, efficient generation becomes increasingly important~\citep{khurana2023natural}. 
However, the inherently sequential nature of  Autoregressive decoding limits throughput and increases latency.
Speculative decoding (SD)~\citep{pmlr-v202-leviathan23a} reduces this sequential bottleneck without changing model outputs: a draft proposes several candidate tokens, and the target verifies them in parallel.
SD speedup depends on both the number of accepted draft tokens and the latency of each draft step.
Accepting more candidates amortizes each target verification over more output tokens, while a slow draft can erase that gain.

\begin{figure}[t]
    \centering
    \includegraphics[width=0.9\linewidth]{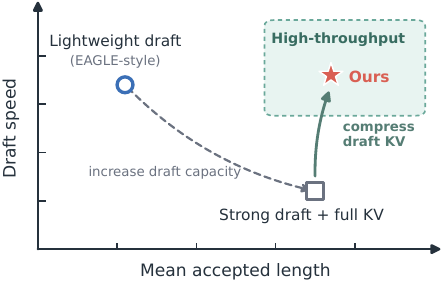}
    \caption{Lightweight drafts are fast but accept fewer tokens; strong full-KV drafts recover acceptance but incur high KV-access cost. Our compressed-memory draft targets high acceptance and speed.}
    \label{fig:pareto_frontier}
\end{figure}

\begin{figure*}[t]
    \centering
    \includegraphics[width=\textwidth]{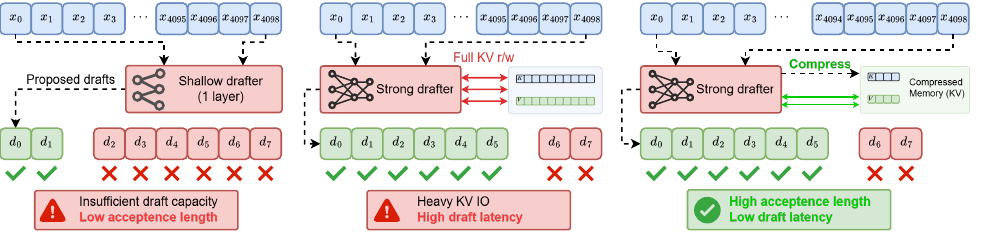}
    \caption{Overview of our proposed framework.}
    \label{fig:overview}
\end{figure*}

When it comes to long-context generation, these applications can condition on prefixes spanning tens of thousands of tokens~\citep{conf/acl/BaiLZL0HDLZHDTL24}.
Longer prefixes increase the cost of each decoding step: attention reads a larger KV cache and processes more key--value pairs, increasing both memory traffic and computation.
Lightweight draft methods such as EAGLE~\citep{pmlr-v235-li24bt} keep draft latency low over short prefixes (Figure~\ref{fig:overview}, left).
Their limited capacity, however, makes it difficult to capture long-range dependencies, so acceptance declines as the context grows~\citep{chen2025rapid,bhendawade-etal-2025-speculative,huo2026repspec}.
Stronger independent drafts recover acceptance (Figure~\ref{fig:overview}, middle), but every draft step streams the full historical KV cache,
whereas KV traffic and computation grows with prefix length.
At long prefixes, this KV cost dominates draft latency and erodes SD's end-to-end speedup.
Figure~\ref{fig:pareto_frontier} summarizes the resulting design target: a draft with sufficient capacity to capture long-range dependencies while maintaining high acceptance and low latency as the prefix grows.

Building on this analysis, we propose \emph{memory-augmented drafting} for long-context SD with independent draft models (Figure~\ref{fig:overview}, right).
Instead of maintaining a full draft KV cache, the draft model keeps a compressed KV memory that carries prediction-relevant information from distant history while preserving exact recent context, thereby preserving both high acceptance and low draft latency (Figure~\ref{fig:pareto_frontier}).
This memory is initialized during the prefill stage---which can run in parallel with target prefill since the draft is an independent model---and is incrementally updated as decoding proceeds.
A lightweight memory adaptor produces the compressed states; only the adaptor parameters are trained while the draft backbone remains frozen, keeping training cost low.
The target verifier retains its unmodified full KV cache, so the standard accept/reject rule applies and speculative decoding remains lossless.
Experiments on Llama~3.1 family targets at prefix lengths up to 32K tokens show that memory-augmented drafting reduces draft-side memory by over 70\% and achieves speedups of up to 2.08$\times$ on the 8B target and 3.33$\times$ on the 70B target, consistently outperforming both lightweight and full-KV draft baselines.

Our contributions are as follows:
\begin{itemize}
    \item We identify the long-context SD dilemma: lightweight drafts lose acceptance as the prefix grows, while strong full-KV independent drafts introduce large overhead.
    
    \item We propose \emph{memory-augmented drafting}, which equips independent draft models with compressed draft-side KV memory that preserves both high acceptance and low draft latency.
    
    \item The design is hardware-friendly: the compressed memory is continuous and append-only, so new slots are materialized and appended without in-place updates to existing entries, making it compatible with efficient KV-cache serving systems.
    
    \item We evaluate on Llama~3.1 family targets across long-input summarization benchmarks at prefix lengths up to 32K tokens, demonstrating consistent speedups over autoregressive decoding and speculative baselines while preserving the lossless guarantee of SD.
\end{itemize}

\section{Related Work}

\subsection{Long-Context Speculative Decoding}

Existing long-context SD methods progressively shift the bottleneck diagnosis from model compute to KV-cache access.
TriForce~\citep{sun2024triforce} and MagicDec~\citep{sadhukhan2025magicdec} drive the draft with StreamingLLM~\citep{xiao2024efficient} attention, dropping middle-range tokens to reduce KV traffic;
TokenSwift~\citep{wu2025tokenswift} uses sparse Medusa heads within an end-to-end optimized decoding system;
QuantSpec~\citep{tiwariquantspec} quantizes the draft KV yet still scans the full prefix at every step;
LongSpec~\citep{yanglongspec} retrains an EAGLE-style draft with attention sinks~\citep{xiao2024efficient}, retaining the shallow-layer capacity bottleneck.
Collectively, these efforts show that compressing the draft-side KV cache is one of the most promising directions for long-context SD, yet none introduces a learnable compressed state to replace distant raw KV.

\subsection{KV Cache Compression}

Methods that compress the KV cache of a single model fall into three families.
\emph{Eviction-based} approaches (StreamingLLM~\citep{xiao2024efficient}, LM-Infinite~\citep{han2024lm}, SnapKV~\citep{li2024snapkv}, AdaKV~\citep{feng2026ada}) discard selected positions, reducing the token count but creating a systematic mismatch when applied asymmetrically between draft and verifier.
\emph{Quantization-based} approaches (KIVI~\citep{liu2023kivi}, ChanMix~\citep{liao2026channel}) trim memory bytes but still scan the full prefix length, and typically require custom GPU kernels for mixed-precision attention.
\emph{Context compression} (Gist~\citep{NEURIPS2023_3d77c6dc}, ICAE~\citep{gecontext}, AutoCompressor~\citep{chevalier2023adapting}, Activation Beacon~\citep{zhanglong}) replaces raw tokens with learnable summary tokens, producing continuous compressed representations that standard attention can consume without specialized kernels.

Among these families, training-aware context compression is most hardware-friendly because its outputs are dense, fixed-shape KV entries compatible with existing serving infrastructure.


\section{Background \& Motivation}
\label{sec:background}

This section revisits the speculative decoding speedup model, then presents two empirical claims that together motivate our design.

\subsection{Speculative Decoding}

Speculative decoding accelerates autoregressive generation through a \textit{draft--verify} paradigm that preserves the target model's output distribution~\citep{pmlr-v202-leviathan23a}. Given a target model $\mathsf{M}_t$ and a cheaper draft model $\mathsf{M}_d$, each iteration first runs $\mathsf{M}_d$ for $L_{\text{draft}}$ autoregressive steps to propose a candidate continuation, and then invokes $\mathsf{M}_t$ in a single parallel forward pass to verify the candidates. Let $L_{\mathrm{acc}}$ denote the mean number of accepted draft tokens per iteration. Standard SD emits these tokens plus one target token: a correction token after the first rejection or a bonus token when all candidates are accepted. Thus, ignoring early termination, the mean number of emitted tokens is $L_{\mathrm{emit}}=L_{\mathrm{acc}}+1$. Let $t_d$ and $t_t$ denote the single-token latency of $\mathsf{M}_d$ and $\mathsf{M}_t$, respectively. The decoding speedup over standard decoding is
\begin{equation}
    \text{Speedup} = \frac{L_{\mathrm{acc}} + 1}{L_{\text{draft}}\cdot \tfrac{t_d}{t_t} + 1}.
    \label{eq:sd_speedup}
\end{equation}
The $+1$ in the numerator counts the extra target token, whereas the $+1$ in the denominator represents one target verification pass normalized by $t_t$. Eq.~\ref{eq:sd_speedup} highlights that practical speedup is governed by two competing factors: a high acceptance length $L_{\mathrm{acc}}$, which requires the draft to faithfully approximate the target, and a small relative draft cost $t_d/t_t$, which requires the draft to remain substantially cheaper than the target.

\begin{figure}[t]
    \centering
    \begin{subfigure}[t]{0.7\linewidth}
        \centering
        \includegraphics[width=\linewidth]{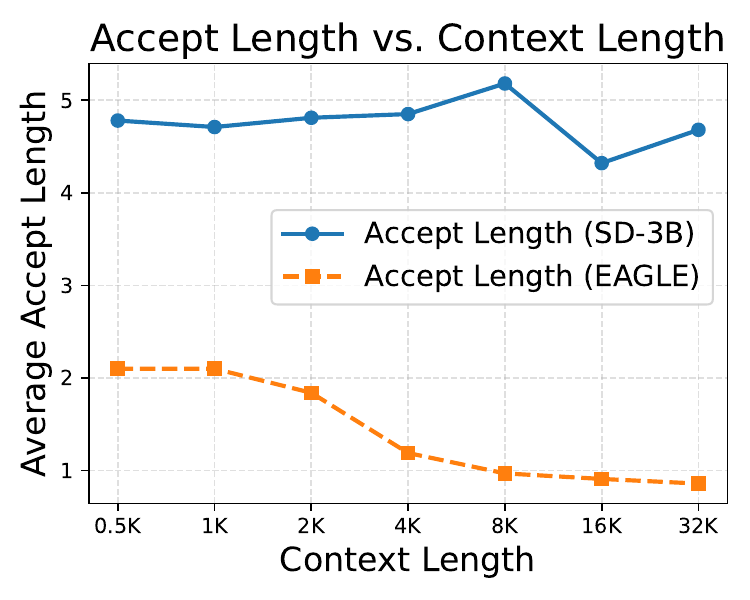}
        \caption{Accepted draft length of EAGLE compared with the 3B draft model for Llama3.1-8B. Both methods use a 10 as draft length.}
        \label{fig:motivation_eagle_throughput_vs_context_length}
    \end{subfigure}
    \hfill
    \begin{subfigure}[t]{0.7\linewidth}
        \centering
        \includegraphics[width=\linewidth]{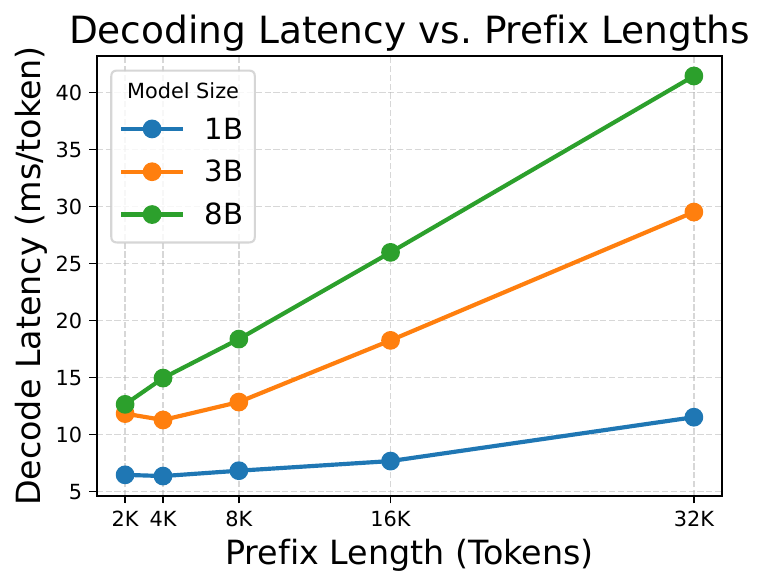}
        \caption{Single token decode latency across prefix lengths and model sizes.}
        \label{fig:motivation_single_token_decode_latency}
    \end{subfigure}

    \caption{Performance and latency analysis across model capacity and context lengths.}
    \label{fig:motivation_combined}
\end{figure}

\subsection{Long-Context SD Needs a Stronger Draft Model}

EAGLE-style methods~\citep{pmlr-v235-li24bt} attach a single lightweight autoregressive layer to the target model. This design satisfies both terms of Eq.~\ref{eq:sd_speedup} on short prefixes: the shallow draft is cheap ($t_d/t_t$ is small) and approximates the target well enough ($L_{\mathrm{acc}}$ is adequate).
Once the prefix grows to tens of thousands of tokens, both properties degrade simultaneously.
Figure~\ref{fig:motivation_eagle_throughput_vs_context_length} shows that EAGLE's $L_{\mathrm{acc}}$ drops monotonically as context length increases for Llama3.1-8B, while the independent Llama3.2-3B draft model maintains a high accepted length.

These experiments show that long-context SD requires capacity: the draft model must be strong enough to approximate the target over long histories. A lightweight single-layer draft cannot meet this requirement.

\subsection{Context Length Dominates Latency}

A stronger independent draft model recovers accepted length, but full-KV drafting requires reading the entire historical KV cache at every draft step. We decompose single-token decode latency into a weight-related term and a KV-access term:
\begin{equation}
    T_{\text{decode}} \;\approx\; \alpha\,d^{2} \;+\; \beta\,L\,d,
    \label{eq:latency_decomp}
\end{equation}
where $d$ is the hidden size and $L$ is the prefix length. The first term reflects loading model parameters and is fixed for a given backbone; the second term reflects streaming the historical KV cache and grows linearly with $L$.
Figure~\ref{fig:motivation_single_token_decode_latency} confirms this decomposition: per-token decode latency scales nearly linearly in $L$, and the slope is dominated by KV access rather than model size.

Hence, the central bottleneck in long-context SD is not draft capacity alone, but the cost of conditioning a strong draft on long histories via full KV-cache access.

\subsection{Summary}

Eqs.~\ref{eq:sd_speedup} and~\ref{eq:latency_decomp} together reshape the design target for long-context SD (Figure~\ref{fig:pareto_frontier}).
Because $L_{\mathrm{acc}}$ scales with draft capacity while $t_d$ is dominated by KV access, an effective long-context draft should retain---or even increase---model capacity to keep $L_{\mathrm{acc}}$ high, yet access a substantially shorter KV memory to keep $t_d/t_t$ low.

\section{Method}

\begin{figure*}
    \centering
    \includegraphics[width=\textwidth]{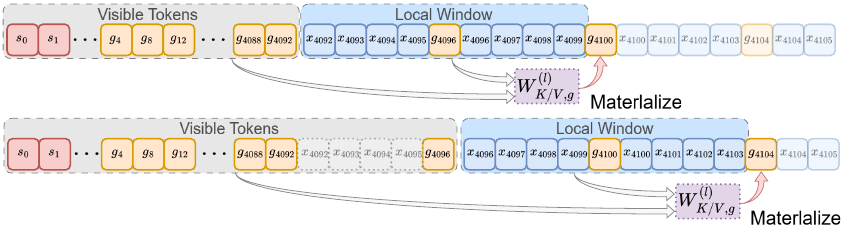}
    \caption{Memory-augmented sliding-window drafting at two consecutive compression boundaries. \textbf{Top:} the draft materializes slot $g_{4100}$ at the current boundary. \textbf{Bottom:} the window has advanced by one slot interval (after 4 decoding steps); the oldest raw KV is evicted, the previously materialized slot is retained as memory, and the next slot $g_{4104}$ is produced at the new boundary.}
    \label{subfig:kv_compressor}
\end{figure*}

Full-KV independent drafting streams the entire historical cache at every step, making draft latency grow linearly with context length; a pure sliding window cuts that cost but removes distant history from the draft, reducing acceptance.
We propose \emph{Memory-Augmented Sliding-Window (MASW) Drafting} (Figure~\ref{subfig:kv_compressor}), which equips an independent draft model with a compact working memory.
MASW is designed to substantially reduce draft latency while preserving the accepted length of a strong independent draft.
The target verifier keeps the standard full KV cache over all original tokens, so speculative decoding remains lossless.
We define the three-part memory in \S\ref{subsec:masw_transition}, describe how slots are materialized in \S\ref{subsec:memory_materialization}, and integrate the design into speculative decoding in \S\ref{subsec:sd_with_masw}.



\subsection{From Full-KV Drafting to Memory-Augmented Sliding Window}
\label{subsec:masw_transition}

An independent draft model that conditions on the full prefix must stream the entire historical KV cache at every draft step, so draft-side latency grows linearly with context length.
The simplest reduction is a sliding window that retains only the most recent raw-token KV entries.
A pure window, however, discards all information beyond the frontier, which misaligns the draft with the target's full-prefix conditionals and hurts acceptance.

We augment the window with \emph{memory slots}: compact states materialized at a fixed rate from completed history and kept in the draft cache after their raw KV is released.
At step $t$, the draft-side working memory is
\begin{equation}
    \mathcal{M}_t
    =
    \mathcal{M}_{\mathrm{sink}}
    \cup
    \mathcal{M}_{\mathrm{local},t}
    \cup
    \mathcal{M}_{\mathrm{slot},t},
    \label{eq:three_part_memory}
\end{equation}
Following the sink-plus-local-window structure of StreamingLLM~\citep{xiao2024efficient}, let $\mathcal{S}=\{1,\ldots,S\}$ denote the first $S$ raw-token positions.
Their KV entries form $\mathcal{M}_{\mathrm{sink}}$ (red, left in Figure~\ref{subfig:kv_compressor}) and remain in the draft cache throughout prefill and decoding.
We call $S=|\mathcal{S}|$ the \emph{sink-token count}.
The local memory $\mathcal{M}_{\mathrm{local},t}$ stores the exact raw KV of the most recent $W$ non-sink tokens (blue, right in Figure~\ref{subfig:kv_compressor}), where $W$ is the \emph{local-window width}:
\begin{equation*}
    \begin{adjustbox}{max width=0.95\linewidth,center}
    $\displaystyle
        \mathcal{M}_{\mathrm{local},t}
        = \bigl\{(K_i,V_i) \mid \max(S{+}1,\, t{-}W{+}1) \le i \le t\bigr\},
    $
    \end{adjustbox}
\end{equation*}
MASW augments this structure with $\mathcal{M}_{\mathrm{slot},t}$, which stores multi-layer KV for memory slots (yellow, interleaved with blue tokens in Figure~\ref{subfig:kv_compressor}) that carry distant context.
During materialization, previous slots carry compressed earlier history into the new slot.
Because the draft accesses only $S$ sink entries, $W$ local entries, and one slot per $r$ tokens, the working-set size scales with $S$, $W$, and $\lfloor t/r \rfloor$ rather than the full prefix length $t$.
The target verifier maintains the usual full KV over all original tokens and is unchanged by this compression.

\subsection{Memory Materialization}
\label{subsec:memory_materialization}

At each compression boundary $\tau_m = mr$, which occurs every $r$ tokens (excluding sink tokens), the draft inserts a memory-slot token $g_m$ (the orange position in Figure~\ref{subfig:kv_compressor}) and runs it through the same transformer blocks as ordinary tokens.
We define $r$ as the \emph{slot interval}: the draft materializes one memory slot for every $r$ raw tokens.
This gives a nominal $r\times$ compression ratio for the materialized history, but not for the complete draft cache, which also retains sink tokens and the local window.
\begin{equation*}
    g_m \leftarrow f_{\theta}\!\bigl(
        \mathcal{M}_{\mathrm{local},\tau_m},\;
        \mathcal{M}_{\mathrm{sink}},\;
        \mathcal{M}_{\mathrm{slot},<m}
    \bigr),
\end{equation*}
where $f_{\theta}$ is the masked forward pass through all transformer layers, and $\theta$ denotes the frozen draft backbone together with the trainable memory-adaptor parameters.
Each slot acts as a \emph{compressed checkpoint} of the current window while reading prior slots, so successive slots form an incremental compression chain rather than disjoint segment summaries.

\paragraph{Structured attention as write policy.}
A structured attention mask, not a hand-designed pooling rule, realizes this update.
After raw tokens are evicted, later tokens can access their information only through the corresponding memory slots.
The next-token objective therefore trains each slot to functionally replace those tokens for subsequent prediction, rather than summarize the document or reconstruct individual KV entries.

\paragraph{Projection and KV materialization.}
At every Transformer layer $l$, memory-slot hidden states use dedicated \emph{mirrored projection matrices} $W_{K,g}^{(l)}$ and $W_{V,g}^{(l)}$ to produce their keys and values:
\begin{equation*}
    \begin{aligned}
        K_r^{(l)} &= H_r^{(l-1)}W_{K,r}^{(l)}, &
        V_r^{(l)} &= H_r^{(l-1)}W_{V,r}^{(l)}, \\
        K_g^{(l)} &= H_g^{(l-1)}W_{K,g}^{(l)}, &
        V_g^{(l)} &= H_g^{(l-1)}W_{V,g}^{(l)}.
    \end{aligned}
\end{equation*}
Only memory-slot states use $W_{\ast,g}^{(l)}$; raw-token states continue to use the original projections $W_{\ast,r}^{(l)}$.
The original projections remain frozen, while the mirrored matrices are initialized from the corresponding backbone weights and constitute the memory adaptor's trainable parameters.
This separate projection branch decouples the information aggregation performed inside the slot hidden state from the KV entries that the slot writes into the cache.
For each layer $l$, the materialized KV pair $(K_{g_m}^{(l)}, V_{g_m}^{(l)})$ is appended to $\mathcal{M}_{\mathrm{slot},t}$ during the draft forward pass.
In the same pass, raw-token KV outside the retained rollback window is evicted; this update does not wait for target verification.
Restricting trainable parameters to $W_{\ast,g}^{(l)}$ keeps the adaptor lightweight and leaves the frozen draft weights untouched, so the draft retains its standalone decoding behavior; we revisit this scope in the Limitations.

\paragraph{Memory update cycle.}
As illustrated in Figure~\ref{subfig:kv_compressor}, the draft model periodically materializes a memory slot every $r$ tokens by compressing the current local window into long-term KV states.
After materialization, raw KV outside the sliding window is discarded while its information remains accessible through the accumulated memory slots.
This materialization-and-eviction cycle repeats throughout decoding.
Each slot $g_m$ shares the RoPE position ID of the immediately following raw token.
The raw token keeps its original position ID, so slot insertion creates a duplicate position ID rather than shifting subsequent raw-token positions.

\subsection{Speculative Decoding with MASW}
\label{subsec:sd_with_masw}

MASW compresses only the draft-side KV; the target verifier retains its full-prefix KV and applies standard parallel verification and rejection sampling.
Thus, memory quality affects accepted length and speed, while target verification preserves the output distribution.

\textbf{Prefill:}
The target model prefills the full prefix to build its complete KV cache.
The draft model processes the same prefix in one forward pass to construct its initial $\mathcal{M}_t$; because the draft is an independent model, its prefill can run in parallel with the target prefill.
This draft-side prefill uses the structured memory mask described in Appendix~\ref{sec:memory-prefill}, so memory slots are materialized during prefix processing without exposing ordinary draft tokens to the full raw-token history.
Slot KV is produced and cached within masked draft forward passes, with no separate retrieval or compression pass.
Prefix slots are created during prefill, and new slots are materialized at compression boundaries during decoding.
Subsequent speculative iterations read only the compact draft memory while verification uses the full cache.

\textbf{Rollback:}
When the target rejects a speculative block, the draft discards speculative local KV and any memory slots created after the last accepted position.
The retained raw-KV rollback window recovers the active local context after rejection.

\section{Evaluation}

We evaluate MASW on long-context speculative decoding tasks spanning two target-model scales (Llama~3.1-8B and 70B) and prefix lengths from 8K to 32K tokens.
\S\ref{subsec:setup} describes the experimental setup; \S\ref{subsec:main_results} reports decoding speedups against autoregressive and speculative baselines; \S\ref{subsec:efficiency_analysis} analyzes draft-side efficiency and the effect of draft prefill on first-token latency; and \S\ref{subsec:ablation} ablates key design choices.

\begin{table*}[t!]
    \centering
    \resizebox{\textwidth}{!}{%
    \begin{tabular}{@{}l|l|ccc|ccc|ccc|ccc@{}}
        \toprule
        & & \multicolumn{3}{c|}{8K} & \multicolumn{3}{c|}{16K} & \multicolumn{3}{c|}{24K} & \multicolumn{3}{c}{32K} \\
        & & Tok./Iter & tok/s & Speedup & Tok./Iter & tok/s & Speedup & Tok./Iter & tok/s & Speedup & Tok./Iter & tok/s & Speedup \\
        \midrule
        \rowcolor[gray]{0.9}\cellcolor{white}
        & AutoRegressive        & 1.00 & 46.15 & 1.00$\times$ & 1.00 & 30.40 & 1.00$\times$ & 1.00 & 22.64 & 1.00$\times$ & 1.00 & 18.21 & 1.00$\times$ \\
        & EAGLE                 & 2.42 & 49.64 & 1.08$\times$ & 1.33 & 17.97 & 0.59$\times$ & 2.36 & 23.75 & 1.05$\times$ & 2.34 & 18.94 & 1.04$\times$ \\
        & EAGLE-3               & 2.07 & 42.46 & 0.92$\times$ & 2.07 & 29.27 & 0.96$\times$ & 2.06 & 22.75 & 1.01$\times$ & 2.06 & 19.24 & 1.06$\times$ \\
        & SWA                   & 2.33 & 42.67 & 0.92$\times$ & 1.49 & 11.32 & 0.37$\times$ & 1.40 & 11.12 & 0.49$\times$ & 1.29 & 9.40  & 0.52$\times$ \\
        & SnapKV                & 3.66 & 41.70 & 0.90$\times$ & 3.08 & 31.21 & 1.03$\times$ & 3.65 & 33.06 & 1.46$\times$ & 2.88 & 23.84 & 1.31$\times$  \\
        & SD L3B                & 5.18 & 54.33 & 1.18$\times$ & 4.32 & 29.94 & 0.98$\times$ & 4.12 & 20.50 & 0.91$\times$ & 4.68 & 17.89 & 0.98$\times$ \\
        \rowcolor[gray]{0.9}\cellcolor{white}
        & Ours 4$\times$ L3B    & 4.79 & 57.42 & 1.24$\times$ & 4.22 & 44.50 & 1.46$\times$ & 4.35 & 41.92 & 1.85$\times$ & \cellcolor{bestspeed}{4.72} & \cellcolor{bestspeed}{37.91} & \cellcolor{bestspeed}{2.08$\times$} \\
        \rowcolor[gray]{0.9}\cellcolor{white} 
        & Ours 8$\times$ L3B    & 3.46 & 41.48 & 0.90$\times$ & 3.47 & 36.61 & 1.20$\times$ & 3.40 & 34.22 & 1.51$\times$ & 3.14 & 26.87 & 1.48$\times$ \\
        \rowcolor[gray]{0.9}\cellcolor{white}
        & Ours 4$\times$ L8B    & 4.94 & 53.64 & 1.16$\times$ & 4.81 & 46.74 & 1.54$\times$ & 4.99 & 42.64 & 1.88$\times$ & 3.64 & 27.62 & 1.52$\times$ \\
        \rowcolor[gray]{0.9}\cellcolor{white} 
        \multirow{-10}{*}{\rotatebox{90}{L3.1-8B}}
        & Ours 8$\times$ L8B    & \cellcolor{bestspeed}{5.51} & \cellcolor{bestspeed}{59.83} & \cellcolor{bestspeed}{1.30$\times$} & \cellcolor{bestspeed}{5.34} & \cellcolor{bestspeed}{51.85} & \cellcolor{bestspeed}{1.71$\times$} & \cellcolor{bestspeed}{5.35} & \cellcolor{bestspeed}{46.59} & \cellcolor{bestspeed}{2.06$\times$} & 4.42 & 35.38 & 1.94$\times$ \\
        \midrule
        \rowcolor[gray]{0.9}\cellcolor{white}
        & AutoRegressive        & 1.00 & 10.09 & 1.00$\times$ & 1.00 & 7.38 & 1.00$\times$ & 1.00 & 5.82 & 1.00$\times$ & 1.00 & 4.79 & 1.00$\times$ \\
        & EAGLE                 & 1.60 & 6.46  & 0.64$\times$ & 1.61 & 5.05 & 0.69$\times$ & 1.47 & 3.89 & 0.67$\times$ & 1.58 & 3.70 & 0.77$\times$ \\
        & EAGLE-3               & 1.26 & 5.08  & 0.50$\times$ & 1.04 & 3.26 & 0.44$\times$ & 1.06 & 2.81 & 0.48$\times$ & 1.06 & 2.48 & 0.52$\times$ \\
        & SWA                   & 3.03 & 5.99  & 0.59$\times$ & 1.86 & 3.43 & 0.47$\times$ & 1.99 & 3.57 & 0.61$\times$ & 1.87 & 3.04 & 0.63$\times$ \\
        & SnapKV                & 4.45 & 9.76  & 0.97$\times$ & 3.84 & 7.76 & 1.05$\times$ & 3.90 & 7.33 & 1.26$\times$ & 3.72 & 6.61 & 1.38$\times$ \\
        & SD L3B                & 4.31 & 24.85 & 2.46$\times$ & 4.14 & 16.97 & 2.30$\times$ & 3.57 & 10.94 & 1.88$\times$ & 4.05 & 9.76 & 2.03$\times$ \\
        & SD L8B                & 4.78 & 22.96 & 2.28$\times$ & 4.65 & 15.59 & 2.11$\times$ & 4.35 & 11.02 & 1.89$\times$ & 4.82 & 9.98  & 2.08$\times$ \\
        \rowcolor[gray]{0.9}\cellcolor{white}      
        & Ours 4$\times$ L3B    & 4.36 & 27.49 & 2.73$\times$ & 3.88 & 19.74 & 2.68$\times$ & 3.60 & 15.53 & 2.67$\times$ & 3.65 & 13.13 & 2.74$\times$ \\
        \rowcolor[gray]{0.9}\cellcolor{white}    
        & Ours 8$\times$ L3B    & 3.72 & 23.46 & 2.33$\times$ & 3.84 & 19.54 & 2.65$\times$ & 3.58 & 16.04 & 2.75$\times$ & 3.43 & 12.32 & 2.57$\times$ \\
        \rowcolor[gray]{0.9}\cellcolor{white}      
        & Ours 4$\times$ L8B    & 4.91 & 29.14 & 2.89$\times$ & \cellcolor{bestspeed}{4.99} & \cellcolor{bestspeed}{24.54} & \cellcolor{bestspeed}{3.33$\times$} & \cellcolor{bestspeed}{4.07} & \cellcolor{bestspeed}{16.93} & \cellcolor{bestspeed}{2.91$\times$} & {3.78} & {13.25} & {2.76$\times$} \\
        \rowcolor[gray]{0.9}\cellcolor{white}    
        \multirow{-11}{*}{\rotatebox{90}{L3.1-70B}}
        & Ours 8$\times$ L8B    & \cellcolor{bestspeed}{5.15} & \cellcolor{bestspeed}{30.54} & \cellcolor{bestspeed}{3.03$\times$} & 4.16 & 20.46 & 2.77$\times$ & 3.78 & 15.73 & 2.70$\times$ & \cellcolor{bestspeed}{3.77} & \cellcolor{bestspeed}{13.54} & \cellcolor{bestspeed}{2.82$\times$} \\
        \bottomrule
    \end{tabular}%
    }
    \caption{Main long-input speculative decoding comparison across two target models and four prefix lengths. \textit{Tok./Iter} denotes the mean number of tokens emitted per speculative iteration and does not apply to AutoRegressive decoding. \textit{tok/s} denotes decoding throughput, and \textit{Speedup} denotes decoding speedup relative to the AutoRegressive baseline. For each target model and prefix length, the best \textit{Speedup} and the corresponding Tok./Iter and tok/s entries are shaded.
    }
    \label{tab:main_results_long_input}
\end{table*}

\subsection{Experimental Setup}
\label{subsec:setup}

\textbf{Models and hardware.}
We use Llama~3.1-8B-Instruct and Llama~3.1-70B-Instruct~\citep{grattafiori2024llama3herdmodels} as target models, and pair them with Llama~3.2-3B-Instruct and Llama~3.1-8B-Instruct as independent draft backbones.
MASW equips each draft backbone with the mirrored projection matrices described in \S\ref{subsec:memory_materialization}; the backbone remains frozen and only the memory-adaptor parameters are trained.
All inference measurements run on a single 8$\times$H100 (80\,GB) node, and the main latency evaluation uses batch size 1.
For heterogeneous batches, per-sequence attention masks disable the extra slot position for sequences that are not at a materialization boundary.
This masking leaves the target KV cache and lossless verification unchanged.

\textbf{MASW configuration.}
We set the local-window width and sink-token count to $W=S=128$, giving every memory slot the same number of visible raw tokens.
MASW uses a 16-token raw-KV rollback window.
We train separate adaptors for nominal 4$\times$ and 8$\times$ compression.

\textbf{Training.}
We train the mirrored projection matrices $W_{\ast,g}^{(l)}$ in two stages.
For pretraining, we sample 2B tokens from RedPajama~\citep{weber2024redpajama} and append an end-of-sequence delimiter to each document.
Supervised fine-tuning (SFT) uses task-oriented long-context data from LongAlpaca~\citep{chen2024longlora} and BookSum~\citep{kryscinski-etal-2022-booksum}, with all sequences truncated to 8K tokens.
The training objective minimizes next-token prediction loss over raw tokens conditioned on the compressed draft memory $\mathcal{D}_t=(\mathcal{M}_{\mathrm{sink}},\mathcal{M}_{\mathrm{local},t},\mathcal{M}_{\mathrm{slot},t})$:
\begin{equation*}
    \mathcal{L}
    =
    \sum_t
    -\log\,
    p(x_t \mid \mathcal{D}_t,\, x_{<t}).
\end{equation*}
Memory-slot positions are excluded from the loss because they serve as compressed carriers rather than prediction targets.

\textbf{Datasets.}
We evaluate on mixed summarization tasks, including GovReport~\citep{huang-etal-2021-efficient}, QMSum~\citep{zhong-etal-2021-qmsum}, and MultiNews~\citep{fabbri-etal-2019-multi} from LongBench-v1~\citep{conf/acl/BaiLZL0HDLZHDTL24}. These tasks require the model to attend to information distributed across the full prefix rather than concentrated at the boundaries.

\textbf{Baselines.}
We compare against four categories of methods:
(1)~vanilla autoregressive (AR) decoding;
(2)~standard speculative decoding (SD) with uncompressed draft KV;
(3)~EAGLE~\citep{pmlr-v235-li24bt} and EAGLE-3~\citep{li2025eagle3scalinginferenceacceleration}, the leading short-context SD methods that use a single lightweight autoregressive layer on top of the target;
(4)~component-level draft-side KV reduction baselines using sliding-window attention (SWA) and SnapKV~\citep{li2024snapkv}.

\textbf{Metrics.}
We report three indicators: the mean number of tokens emitted per speculative iteration (\textit{Tok./Iter}), decoding throughput (\textit{tok/s}), and decoding speedup over the AR baseline (\textit{Speedup}).
All main experiments and ablations use greedy decoding with temperature $T{=}0$.
Appendix~\ref{sec:detailed-experimental-setting} provides table-specific configurations, detailed baseline implementations, and evaluation protocols.

\subsection{Main Results}
\label{subsec:main_results}

Table~\ref{tab:main_results_long_input} compares all methods across two target scales and prefix lengths from 8K up to 32K.
EAGLE and EAGLE-3 provide little or inconsistent acceleration at long contexts, confirming the capacity bottleneck identified in \S\ref{sec:background}.
SWA is particularly fragile because discarding distant KV sharply reduces accepted length.
Full-KV SD also loses speedup as the prefix grows because draft-side KV traffic increases with context length.
MASW avoids this decline.
Its strongest settings gain speedup with context on the 8B target and remain consistently strong on the 70B target, where the larger verifier amortizes the compressed draft cost.
Across draft backbones, 4$\times$ compression tends to preserve more tokens per iteration, whereas 8$\times$ further reduces draft latency, making the best ratio dependent on target scale and prefix length.

\subsection{Effectiveness \& Efficiency Analysis}
\label{subsec:efficiency_analysis}
\label{subsec:performance_analysis}

We analyze MASW's draft-side effectiveness and efficiency in terms of prefill latency, decoding latency, and peak memory usage.

\paragraph{Prefill latency.}
MASW materializes prefix memory during draft-side prefill, which adds computation.
However, Table~\ref{tab:prefill_latency} shows that the memory-augmented 8B draft still prefills faster than the 70B target at all tested prefix lengths.

\begin{table}[H]
    \centering
    \resizebox{\linewidth}{!}{%
    \begin{tabular}{@{}lrrr@{}}
        \toprule
        Model & 8K & 16K & 32K \\
        \midrule
        8B draft, original & 306.5 & 711.2 & 1835.2 \\
        8B draft, with memory & 751.6 & 2350.2 & 7992.5 \\
        70B target, original & 2212.9 & 5480.3 & 12316.4 \\
        \bottomrule
    \end{tabular}}
    \caption{Prefill latency (ms) across prefix lengths.}
    \label{tab:prefill_latency}
\end{table}

Let $P_d$ and $P_t$ denote draft and target prefill latency, respectively.
Because both prefills run concurrently,
\begin{equation}
    T_{\mathrm{prefill}}^{\mathrm{MASW}} = \max(P_d,P_t) = P_t \quad (P_d<P_t).
\end{equation}
Thus, the additional work is hidden behind target prefill and does not increase TTFT in this setting.

\paragraph{Draft-side efficiency.}

\begin{table}[h]
    \centering
    \resizebox{0.95\linewidth}{!}{%
    \begin{tabular}{@{}l|rrcc@{}}
        \toprule
        Draft     & Weights                     & 32K Extra Peak   & Tok./Iter & Latency (ms) \\
        \midrule
        3B         & \multirow{3}{*}{5.98 GB}  & 17.21 GB & 4.05 & 41.34 \\
        Ours 3B 4$\times$ &                    & 4.42 GB  & 3.65 & 13.94 \\
        Ours 3B 8$\times$ &                    & 3.98 GB  & 3.43 & 12.39 \\ \midrule
        8B         & \multirow{3}{*}{14.95 GB} & 18.02 GB & 4.82 & 54.92 \\
        Ours 8B 4$\times$ &                    & 5.05 GB  & 3.78 & 15.37 \\
        Ours 8B 8$\times$ &                    & 4.55 GB  & 3.77 & 14.02 \\
        \bottomrule
    \end{tabular}}
    \caption{Draft-side effectiveness and efficiency under the L3.1-70B target at 32K prefix length. \textit{Weights} reports frozen draft-backbone weight memory. \textit{Extra Peak} reports all peak GPU memory beyond the backbone weights, including memory-adaptor weights when present, the KV cache, attention masks, logits, and temporary buffers.}
    \label{tab:effectiveness_efficiency_analysis}
\end{table}

Table~\ref{tab:effectiveness_efficiency_analysis} compares full-KV and MASW drafting at a 32K prefix under the L3.1-70B target.
MASW reduces extra peak memory from 17.21--18.02 GB to 3.98--5.05 GB and draft latency from 41.34--54.92 ms to 12.39--15.37 ms.
Meanwhile, Tok./Iter decreases only from 4.05 to 3.43 for the 3B draft and from 4.82 to approximately 3.78 for the 8B draft.
Increasing compression from 4$\times$ to 8$\times$ provides further memory and latency savings while preserving Tok./Iter on the 8B draft.

\subsection{Ablation Study}
\label{subsec:ablation}

We ablate four design choices in MASW: local-window size, mirrored-projection initialization, training procedure, and training-context length.

\subsubsection{Local-Window Size}

\begin{figure}[H]
    \centering
    \includegraphics[width=0.7\linewidth]{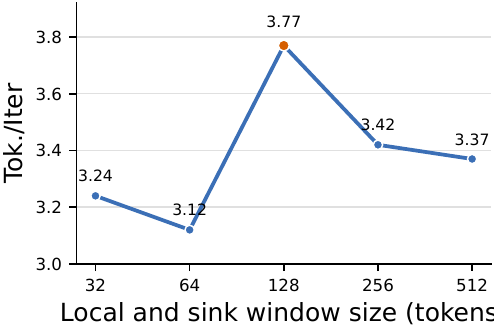}
    \caption{Window-size ablation.}
    \label{fig:window_size_ablation}
\end{figure}

We vary the local-window size from 32 to 512 tokens for the L3.1-8B draft at nominal 8$\times$ compression, using the L3.1-70B target and a 32K prefix.
For each configuration, we set the sink-token count equal to the local-window width ($S=W$) so that every memory slot observes the same number of raw local tokens.
Figure~\ref{fig:window_size_ablation} shows that accepted length remains relatively stable across the tested range and peaks at 3.77 with a 128-token window.
We therefore use $W=S=128$ in the main experiments.

\subsubsection{Mirrored Projection Initialization}

We compare two strategies for initializing the mirrored projection matrices $W_{\ast,g}^{(l)}$: random initialization versus copying from the backbone's pretrained KV projections $W_{\ast,r}^{(l)}$.
We pretrain the L3.2-3B memory adaptor at nominal 4$\times$ compression under each strategy.

\begin{figure}

    \centering
    \begin{subfigure}[t]{0.7\linewidth}
        \centering
        \includegraphics[width=\linewidth]{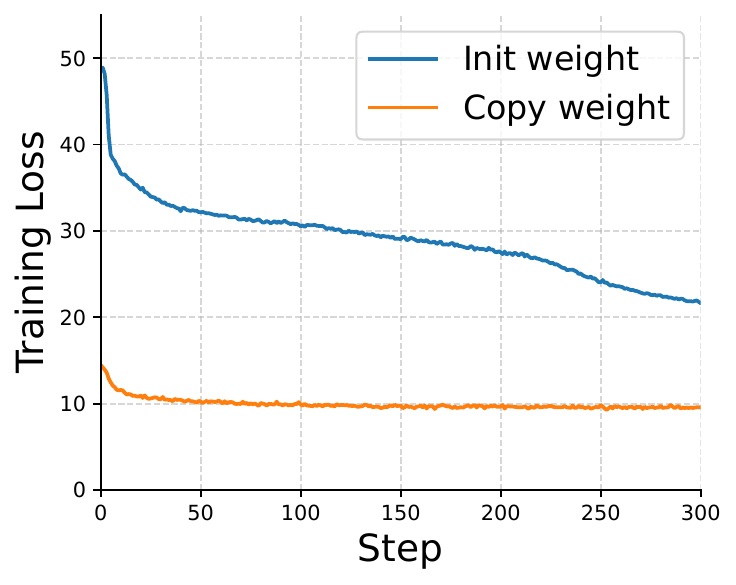}
        \caption{Pretraining loss.}
        \label{fig:model_weight_init_loss}
    \end{subfigure}
    \hfill
    \begin{subfigure}[t]{0.7\linewidth}
        \centering
        \includegraphics[width=\linewidth]{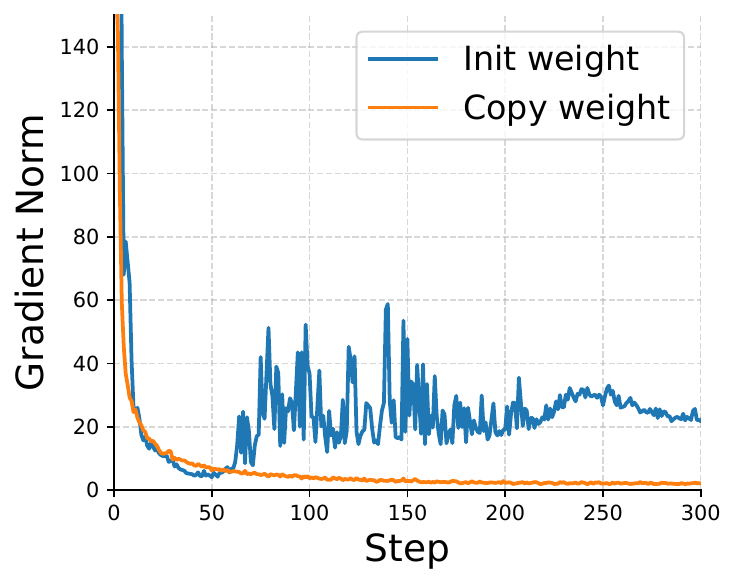}
        \caption{Gradient norm.}
        \label{fig:model_weight_init_grad_norm}
    \end{subfigure}
    \caption{Pretraining loss and gradient norm for the L3.2-3B memory adaptor at nominal 4$\times$ compression.
    }
    \label{fig:model_weight_init}
\end{figure}

Figure~\ref{fig:model_weight_init} shows a persistent gap between the two strategies.
Random initialization starts at a much higher loss and, even after 300 steps, never reaches the range that copy-weight initialization already occupies at step 0.
The gradient norm tells the same story: random initialization stays elevated and noisy throughout training, whereas copy-weight initialization decays into a small, stable regime within the first tens of steps.
Preserving the backbone's KV geometry gives the optimizer a strong starting point, so we adopt copy-weight initialization as the default throughout all experiments.
\subsubsection{Training Recipe}

We compare three training recipes for the memory adaptor: continued \emph{pretraining} alone, \emph{SFT} alone, and the two-stage \emph{pretrain-then-SFT} pipeline.
Table~\ref{tab:training_procedure} reports mean Tok./Iter at 16K context across both draft backbones and nominal compression ratios.

\begin{table}[t]
    \centering
    \begin{adjustbox}{max width=0.85\linewidth}
    \begin{tabular}{@{}lcccc@{}}
        \toprule
        \textbf{Model} & \textbf{Ratio} & PT only & SFT only & PT + SFT \\
        \midrule
        \multirow{2}{*}{L3B}
            & 4$\times$ & 3.70 & 3.23 & \textbf{4.15} \\
            & 8$\times$ & 3.78 & 3.24 & \textbf{4.78} \\
        \midrule
        \multirow{2}{*}{L8B}
            & 4$\times$ & 4.22 & 3.18 & \textbf{4.81} \\
            & 8$\times$ & 3.87 & 3.01 & \textbf{5.34} \\
        \bottomrule
    \end{tabular}
    \end{adjustbox}
    \caption{Mean number of tokens emitted per speculative iteration (Tok./Iter) at 16K context under three training recipes. \textbf{Bold} marks the best recipe per row.}
    \label{tab:training_procedure}
\end{table}

PT + SFT achieves the highest Tok./Iter across all settings.
SFT alone is consistently weaker, suggesting that supervised data cannot by itself teach the adaptor to form robust compressed representations.
Pretraining provides a stronger starting point, but still needs SFT to align the memory slots with the downstream drafting distribution.
The advantage of the two-stage recipe becomes more pronounced under nominal 8$\times$ compression, where the adaptor must preserve more information in fewer slots.
We therefore use pretrain-then-SFT as the default recipe.

\subsubsection{Training-Context Length}

\begin{table}[h]
    \centering
    \begin{adjustbox}{max width=0.85\linewidth}
    \begin{tabular}{@{}lcccc@{}}
        \toprule
        \textbf{Training Context} & \textbf{8K} & \textbf{16K} & \textbf{24K} & \textbf{32K} \\
        \midrule
        8K  & \textbf{5.15} & \textbf{4.16} & \textbf{3.78} & \textbf{3.77} \\
        32K & 4.96 & 4.01 & 3.45 & 3.58 \\
        \bottomrule
    \end{tabular}
    \end{adjustbox}
    \caption{Mean number of tokens emitted per speculative iteration (Tok./Iter) for L3.1-8B adaptors trained with 8K or 32K contexts at nominal 8$\times$ compression.}
    \label{tab:training_context_length}
\end{table}

We train the L3.1-8B adaptor at nominal 8$\times$ compression with either 8K- or 32K-token sequences, using a fixed 2B-token training budget, and evaluate both adaptors with the L3.1-70B target.
As shown in Table~\ref{tab:training_context_length}, the adaptor trained on 8K sequences achieves higher Tok./Iter at every evaluation length, including 32K.
Rather than learning a representation tied to an absolute context length, MASW learns a relative slot-materialization operation: each slot combines the bounded local window, sink tokens, and earlier slots, and the same operation is repeatedly applied as the context grows.
Under a fixed token budget, 8K sequences provide more training instances, which likely improves the data efficiency of learning this operation.
These results show that the learned compression extrapolates beyond the training context length and that longer training sequences are not necessary in this setting.

\section{Conclusion}

We introduced Memory-Augmented Sliding-Window (MASW) Drafting, which replaces the full draft KV with a compact working memory of sink tokens, an exact local window, and periodically materialized memory slots produced by trainable mirrored projection matrices.
Across long-context settings, this design substantially reduces draft-side memory and latency while preserving much of the acceptance quality of full-KV drafting.
Because the target verifier operates on the unmodified full KV cache, the lossless guarantee of speculative decoding is preserved.
Lower latency and serving cost may also facilitate undesirable uses of existing language models, such as large-scale generation of misleading, abusive, or otherwise harmful content.

\section*{Limitations}

Because training resources are limited, we train the memory adaptor on only 2B tokens with an 8K context length. This leaves open a broader study of adaptor-training configurations, including longer contexts, larger training corpora, and full fine-tuning of the draft model so that it can learn context compression more natively.
Training only the mirrored K/V projections keeps the adaptor lightweight, yet may cap how much information each slot can encode; jointly fine-tuning a larger subset of draft parameters, such as MLP blocks or low-rank updates on the backbone, is a natural extension we leave to future work.
MASW currently fixes the local-window size, sink-token count, and slot interval to obtain regular masks, stable cache layouts, and bounded materialization overhead; adaptive allocation remains future work.
MASW could also compress the draft path in self-speculative decoding. However, using the target model or part of it for long-context drafting may weaken cost-effectiveness; we leave this trade-off to future work.

\section*{Acknowledgments}

This work was supported by the National Natural Science Foundation of China (NSFC) under Grant No. 62306256 and the Natural Science Foundation of Guangdong Province under Grant No. 2025A1515010261.

\paragraph{AI Assistance}

We used AI assistants for preliminary information retrieval and organization, code debugging, and grammar-level checking and polishing.
All retrieved information, suggested references, code changes, and textual edits were manually checked and verified by the authors.

\newpage
\bibliography{ref/reference}

\newpage
\appendix


\section{Scientific Artifacts}
\label{sec:scientific-artifacts}

This work uses existing scientific artifacts, including publicly available research models and datasets, solely for research and evaluation purposes. We cite the original creators of the models, datasets, baselines, and software frameworks used in our experiments in the corresponding experimental setup and baseline sections, as well as in the references. We use these artifacts consistently with their stated intended uses and access conditions, and we follow the licenses or terms specified by their original providers.

We do not collect new human-subject data, annotate private data, or redistribute raw datasets. The datasets used in our experiments are standard public research benchmarks. To the best of our knowledge, based on their original documentation, these datasets are not intended to contain personally identifying information. Since our work focuses on efficiency evaluation rather than dataset construction, we rely on documentation from the original artifact creators for details such as data source, domain, language coverage, and usage restrictions. We also document the models, datasets, evaluation settings, and implementation details used in our experiments.

\section{Potential Risks}

MASW accelerates inference but does not add new content-generation capabilities, and it inherits the safety, bias, and privacy risks of its underlying models and training data.
Greater inference efficiency may nevertheless enable harmful uses of existing models at a larger scale.
The lossless guarantee also relies on verifying every draft proposal with the target model; deployments that weaken or bypass this verification may change model outputs.
Practitioners should therefore retain full target verification and apply the same access controls, monitoring, and content safeguards required for the underlying model.

\newpage
\section{Memory Prefill}
\label{sec:memory-prefill}

\begin{figure}[h]
    \centering
    \includegraphics[width=0.85\linewidth]{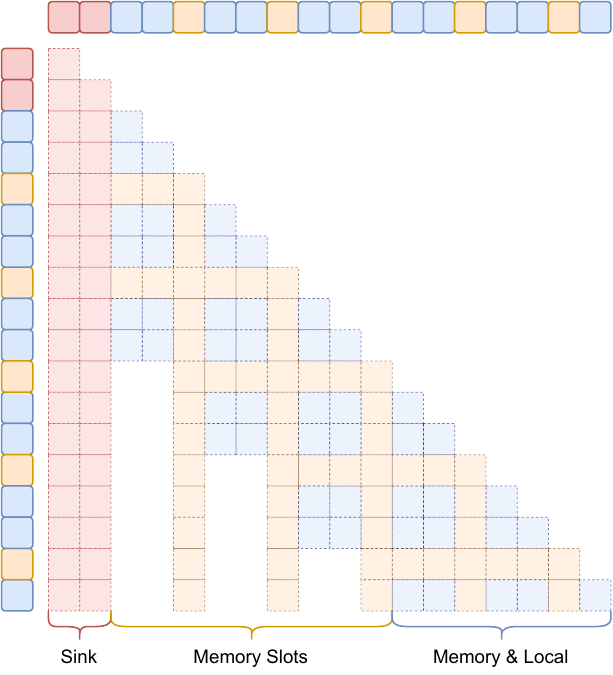}
    \caption{Memory prefill attention mask.}
    \label{fig:memory_prefill}
\end{figure}

During draft-side prefill, the attention mask enforces the same memory-write policy used during decoding.
Sink tokens remain globally visible to stabilize attention, and each memory-slot token can attend to previously materialized slots together with the raw tokens in its assigned local unit and neighboring boundary tokens.
Ordinary raw tokens, in contrast, are restricted to the compact memory view: sink tokens, available memory slots, and the bounded local window.
This asymmetric visibility lets the prefill pass build the initial compressed memory $\mathcal{M}_t$ while avoiding a full raw-token KV cache for the draft model.
Figure~\ref{fig:memory_prefill} illustrates this mask structure.

\section{Detailed Experimental Setting}
\label{sec:detailed-experimental-setting}




\paragraph{Component-level baselines.}
We reproduce SWA and SnapKV within the same Hugging Face Transformers framework~\citep{wolf-etal-2020-transformers}.
For both baselines, the draft and target use the same model backbone and weights.
The draft generates tokens autoregressively using SWA or SnapKV for draft-side KV reduction, whereas verification uses the full target KV cache.
SWA uses a sliding-window size of 1,024 tokens.
SnapKV retains 4,096 KV entries and otherwise follows its official default configuration.
For EAGLE and EAGLE-3, we use chain-based proposal generation with the released checkpoints and the remaining official inference settings.
All speculative methods in Table~\ref{tab:main_results_long_input} propose five draft tokens per iteration; the separate motivation experiment in Figure~\ref{fig:motivation_eagle_throughput_vs_context_length} uses a 10-token proposal horizon.

\paragraph{Decoding and timing.}
The reported 8K, 16K, 24K, and 32K lengths refer to the input prefix before generation and exclude output tokens.
We filter examples from GovReport, QMSum, and MultiNews whose tokenized inputs match the evaluated prefix length.
Each run generates at most 1,024 output tokens and stops early when the model emits an end-of-sequence token.
\textit{Tok./Iter} denotes the mean number of tokens emitted per speculative iteration.
Decoding throughput (\textit{tok/s}) is the number of generated output tokens divided by decoding wall-clock time; draft and target prefill are excluded.
\textit{Speedup} denotes decoding speedup over the corresponding AutoRegressive baseline under the same prefix length and timing scope.

\FloatBarrier







\section{Notations}

\begin{table*}[h]
    \centering
    \resizebox{\textwidth}{!}{%
    \begin{tabular}{@{}lll@{}}
        \toprule
        Part & Notation & Explanation \\
        \midrule
        Speculative decoding & $\mathsf{M}_d$, $\mathsf{M}_t$ & draft and target models \\
                              & $L_{\text{draft}}$ & number of draft tokens proposed per iteration; fixed to 5 \\
                              & $L_{\mathrm{acc}}$ & mean number of accepted draft tokens per iteration \\
                              & $L_{\mathrm{emit}}$ & mean emitted tokens per nonterminal iteration; reported as Tok./Iter \\
                              & $t_d$, $t_t$ & single-token latency of the draft and target models \\
        \midrule
        Draft memory & $\mathcal{D}_t$ & compact draft-side working memory at step $t$ \\
                     & $\mathcal{M}_{\mathrm{sink}}$ & exact KV for the first $S$ raw tokens \\
                     & $\mathcal{M}_{\mathrm{local},t}$ & exact raw-token KV in the local window \\
                     & $\mathcal{M}_{\mathrm{slot},t}$ & materialized KV entries for memory slots \\
                     & $\mathcal{S}=\{1,\ldots,S\}$ & sink-token positions \\
                     & $S$ & sink-token count \\
                     & $W$ & local-window width \\
                     & $r$ & slot interval; one slot per $r$ raw tokens \\
        \midrule
        Memory slots & $g_m$ & memory-slot token for unit $m$ \\
                     & $\tau_m=mr$ & compression boundary for slot $g_m$ \\
                     & $\mathcal{A}(g_m)$ & write scope of memory-slot token $g_m$ \\
                     & $\mathcal{G}_{<m}$, $U_m$, $\mathcal{N}_m$ & earlier slots, current unit, and neighboring tokens \\
        \midrule
        Transformer and KV & $d$ & hidden size \\
                           & $L$ & prefix length in the latency model \\
                           & $K_r^{(l)}$, $V_r^{(l)}$ & raw-token key and value at layer $l$ \\
                           & $K_g^{(l)}$, $V_g^{(l)}$ & memory-slot key and value at layer $l$ \\
                           & $W_{K,r}^{(l)}$, $W_{V,r}^{(l)}$ & frozen backbone KV projections for raw tokens \\
                           & $W_{K,g}^{(l)}$, $W_{V,g}^{(l)}$ & trainable mirrored KV projections for memory slots \\
                           & $W_{\ast,r}^{(l)}$, $W_{\ast,g}^{(l)}$ & shorthand for raw-token and memory-slot KV projections \\
        \bottomrule
    \end{tabular}
    }
    \caption{List of notation used in this paper.}
    \label{tab:notation}
\end{table*}

\end{document}